# Empirical Study of Named Entity Recognition Performance Using Distribution-aware Word Embedding

Xin Chen (xinc4)
xinc4@illinois.edu
ISE Department, University of Illinois at Urbana-Champaign

Qi Zhao (qiz2)
qiz2@illinois.edu
ISE Department, University of Illinois at Urbana-Champaign

Xinyang Liu (xl50)
xl50@illinois.edu
ISE Department, University of Illinois at Urbana-Champaign

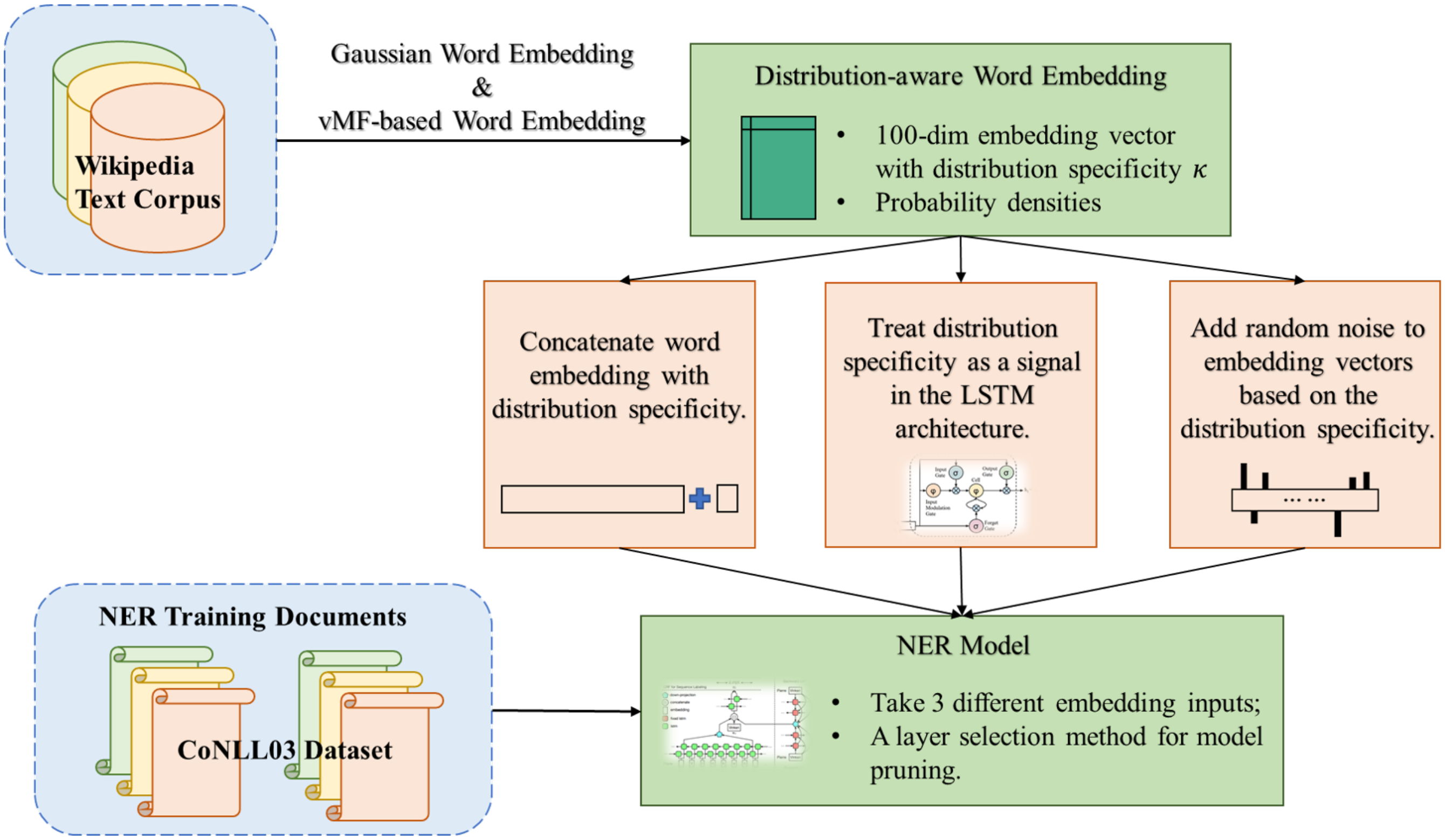

**Figure 1: Framework of NER with Distribution-aware Word Embedding**

## ABSTRACT

With the fast development of Deep Learning techniques, Named Entity Recognition (NER) is becoming more and more important in the information extraction task. The greatest difficulty that the NER task faces is to keep the detectability even when types of NE and documents are unfamiliar. Realizing that the specificity information may contain potential meanings of a word and generate semantic-related features for word embedding, we develop a distribution-aware word embedding and implement three different methods to make use of the distribution information in a NER framework. And the result shows that the performance of NER will be improved if the word specificity is incorporated to existing NER methods.

## KEYWORDS

named entity recognition, distribution-aware, robustness

## 1 INTRODUCTION

Modern Named Entity Recognition (NER) models perform remarkably well on standard English datasets while the performance would drop greatly when casing information is missing and when a surface name has multiple meanings or is marked as multiple entities in a corpus. [12]. Most of the current work on named entity recognition would apply a fixed word embedding to train the model, which will result in a NER model with less flexibility and comprehensiveness. Therefore, we would like to test the performance of adopting distribution-aware word embedding methods like [2] and [23] in tackling NER tasks. Rather than using a fixed vector as the word embedding, variance information of the word embedding can be used in the NER model.

There have been different ways to capture the word distributional specificity as defined in [13]. The paper assumes there is a scalar $\kappa_w \geq 0$ correlated with each word $w$ indicating how specific the word meaning is. And [14] assumes that the generation of a word is dependent on its belonging paragraph with a von Mishes-Fisher (vMF) distribution which is parameterized by a mean vector $\boldsymbol{\mu}$ and a concentration parameter $\kappa$ in the spherical space. Such a parameter indicating word distributional specificity may contain potential meanings of a word and generate semantic-related features for training NER models.

In this project, we aim to study how the performance of NER models can be improved by adopting the distribution-aware word embedding and explore ways to make good use of the word distributional specificity. Then the report is organized as follows. Section 2 reviews the related work and state-of-the-art methods in developing distribution-aware word embeddings and improving NER performances. Next, Section 3 introduces how we developed our own word embeddings and three ways to make use of the distributional specificity information. And then experiments and result analysis are presented in Section 4. Finally, Section 5 concludes the project work and provides our ideas for future work.

# 2 RELATED WORK

This project involves the advanced word embedding methods and efforts in improving the robustness of NER models, so we summarize the related work from these two aspects below.

## 2.1 Word Embedding

Distributed representation represents words in low dimensional real-valued dense vectors where each dimension stands for a latent feature. Word Embedding is an example of distributed representation used in Deep Learning for NER models and is typically pretrained over large collections of text via unsupervised algorithms. Commonly used models include Word2Vec [15] and GloVe [16] which train each word as a vector in the Euclidean space.

Apart from common embedding methods which map words to estimated fixed vectors, Luke Vilnis et al. first propose the idea of mapping words to probability densities in their paper [23]. They present a method for learning representations in the space of Gaussian distributions so that the mapping can better capture uncertainty about representation and its relationship. The embedding and variance are learned by minimizing the KL divergence of distribution between the center word and its context words. They demonstrate the effectiveness of their embedding on a linguistic similarity task requiring asymmetric comparisons.

Recently spherical word embedding [13] has been proposed to resolve the inconsistency between the training of word embedding in Euclidean space and usage in spherical space which may lead to sub-optimal solutions. They find the probability relation between the center word and its context words follows vMF distribution in spherical space after adding one $\kappa$. Thus the model can learn both word embedding and word distributional specificity that maximize the probability of the context embedding generated by the center word's vMF distribution.

## 2.2 Named Entity Recognition

Named Entity Recognition is the task of identifying named entities in text. Mathematically, given a sequence of tokens $s = < w_1, w_2, ..., w_N >$, NER is to output a list of tags $< I_s, I_e, t >$, each of which is a named entity of tokens in the sequence s. Here, $I_s$ and $I_e$ are the start and the end of a named entity mention; $t$ is the entity type from a predefined category set. NER has many applications in the field of natural language processing, including machine translation, automatic summarization, internet search engines, semantic annotation, automatic question-answering, information retrieval, etc. Recent deep neural network architectures (e.g. [8] and [21]) for NER have achieved state-of-the-art performance with minimal feature engineering.

Since the first shared task on NER, many shared tasks and datasets for NER have been created [25]. CoNLL 2002 [19] and CoNLL 2003 [18] were created from newswire articles in four different languages (Spanish, Dutch, English, and German) and focused on 4 entities: PER (person), LOC (location), ORG (organization) and MISC (miscellaneous including all other types of entities). Besides these typical datasets, different NER tasks have also been developed in other domains (e.g. the biomedical domain [7], [20] and social media [1], [10]).

With respect to the learning process, NER has developed from knowledge-based systems to unsupervised and bootstrapped systems and then feature-engineered supervised systems. Knowledge-based NER systems don't require annotated training data as they rely on lexicon resources and domain-specific knowledge. Precision is generally high for knowledge-based NER systems because of the lexicons while recall is often low due to domain and language-specific rules and incomplete dictionaries. For the feature-engineered supervised systems, Hidden Markov Models (HMM), Support Vector Machines (SVM), Conditional Random Fields (CRF), and decision trees were common machine learning systems for NER. Different models have also exploited word-level architecture [4], character-level architecture [17] or a mixture of these two [11]. An example of word-level architecture and character-level architecture is shown in Figure 2 and Figure 3 separately.

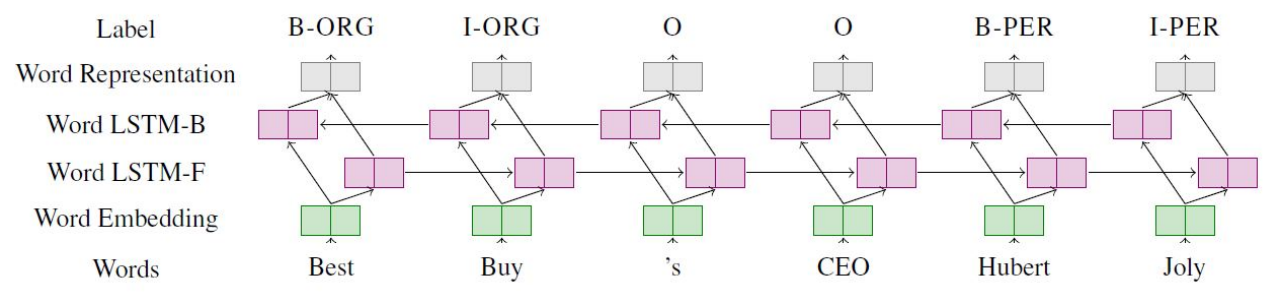

**Figure 2: Word Level NN Architecture for NER**

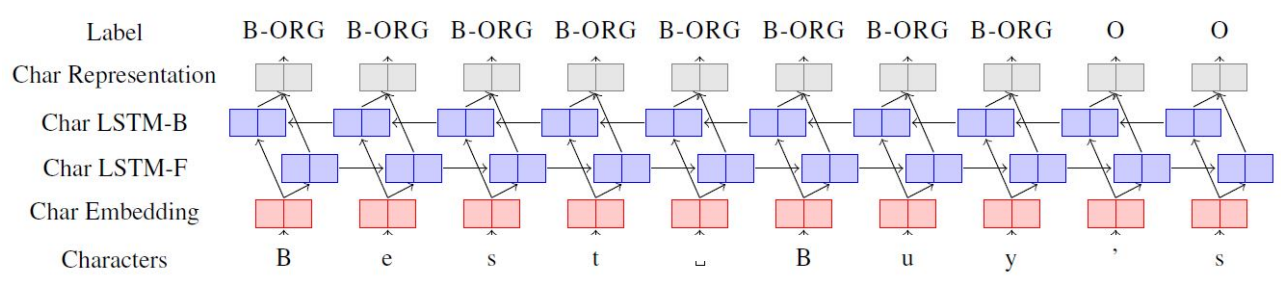

**Figure 3: Character Level NN Architecture for NER**

# 3 METHODOLOGY

## 3.1 Overview

In this section, we will explain how we develop the distribution-aware word embedding (Section 3.2) and how it is adopted in the selected NER model (Section 3.3 - 3.5, each of which introduces one possible manner). The framework of this project is shown in Figure 1. We first train Gauss-based embedding and vMF-based word embedding with their distribution specificity and then modify the NER model in [9] to make use of word specificity information and test the performance.

## 3.2 Distribution-aware Word Embedding

Inspired by [13], [14], we train our word embedding directly in the spherical space, which is validated to have high efficiency and state-of-the-art performances on various text embedding tasks. Using the negative log-likelihood as our objective function, we have

$$L = \sum_{w_{i+j} \in d, -h \leq j \leq h, j \neq 0} log(p(w_{i+j}|w_i)) \tag{1}$$

where w is one word, d is the current document, i refers to the center word, j refers to the context word and h is the size of local context window.

Then we define the probability expression in the above equation as

$$p\left(w_{i+j}|w_i\right) = \frac{\exp\left(\kappa_{w_i} \boldsymbol{u}_{w_i}^\top \boldsymbol{v}_{w_{i+j}}\right)}{\sum_{w' \in V} \exp\left(\kappa_{w_i} \boldsymbol{u}_{w_i}^\top \boldsymbol{v}_{w'}\right)} \tag{2}$$

the contexts vectors are assumed to be generated from the vMF distribution with a mean vector $\boldsymbol{\mu}$ and a concentration parameter $\kappa$. The model learns both word embedding and word distributional specificity by maximizinng the probability of the context vectors getting generated by the center word's vMF distribution[13].

As introduced before, the concentration parameter $\kappa$ stands for the variance of word embedding, which captures different semantic meanings of words in different contexts. We use Wikipedia text corpus containing 2.4 billion tokens [1] to train our word embedding. Apart from the distribution-aware word embedding, we also use the GloVe method [16] to train another word embedding as a benchmark and the Gauss-based embedding as a comparison. Then we perform word analogy and word similarity tasks to validate the effectiveness of the word embedding that we have trained.

## 3.3 Method 1: Concatenated Word Embedding

The first attempt to make use of the distribution specificity is to concatenate the pre-trained $\kappa$ with each 100-dim embedding vector. By reforming the input embedding as a 101-dim vector, the specificity information can be directly fed into the existing NER model. Considering the dropout rate in the NER model is high up to 0.5, the distribution specificity information with one dimension is easily neglected in the training process. We also conduct experiments to see whether increasing the dimensions of the distribution specificity will actually let the information be recognized and thus improve the performance. The formula of the new embedding for a word $i$: $\boldsymbol{w}_i'$ is represented in equation (3), in which $\boldsymbol{w}_i$ is the original word embedding vector, $\boldsymbol{\kappa} \in \mathcal{R}^n$ is the vector of distribution specificity and $n$ is the number of replications of the value. In the experiment, we may choose different values for $n$ to make the information from distribution specificity recognized by the NER model.

$$\boldsymbol{w}_i' = \boldsymbol{w}_i \oplus \boldsymbol{\kappa} \tag{3}$$

## 3.4 Method 2: Gate Mechanism

Inspired by the work of Ying et al. [8] which uses a reliability signal to dynamically select features from word-level embedding and character-level embedding via a gate mechanism, we use the pre-trained $\kappa$ as a specificity signal to control the feature composition between word-level embedding and character-level embedding. We calculate the final representation of word i as

$$\boldsymbol{x}_i = \boldsymbol{w}_i * \boldsymbol{gate} \oplus \boldsymbol{c}_i \tag{4}$$

where $\boldsymbol{w}_i$ is word-level representation, $\boldsymbol{c}_i$ is character-level representation, $\boldsymbol{gate} = sigmoid(\kappa * \boldsymbol{w} + b)$ , $*$ is Hadamard product operator, $\oplus$ is concatenation operator, w (100-dim) and b (1-dim) are parameters of the gate which will be updated during the training process. They are initialized to zeros. For words with larger variance, the embedding is not as reliable as words with smaller variance thus the model will rely less on the word-level embedding and try to learn its meaning either through character-level embedding or context word embedding.

## 3.5 Method 3: Embedding with Random Sampling

Since we wish to capture distributional information with the embedding, one attempt is to encode $\kappa$ in a random sampling noise and add it to the original embedding. The intuition behind this is to assume that the embedding is a random variable rather than a deterministic value. By incorporating random noise in the embedding we are actually sampling the embedding from a random variable. What we aim to achieve is that first by sampling we can reduce overfitting in the training dataset. More importantly by introducing the distribution by sampling, what we wish to achieve is a more robust model in the sense that it is less sensitive to the various distribution of different test sets so as to lead to a better generalization result.

Considering the distribution to sample from, there will be two choices: vMF distribution and normal distribution.

The intuitive idea is to choose vMF distribution since the embedding is trained from vMF distribution and the natural solution is to again sample from this distribution. However, if we observe the density of vMF distribution:

$$f_p(\boldsymbol{x}; \boldsymbol{\mu}, \kappa) = C_p(\kappa) exp(\kappa \boldsymbol{\mu}^T \boldsymbol{x}) \tag{5}$$

with $p$ being the dimension, here $p = 100$. And

$$C_p = \frac{\kappa^{p/2-1}}{(2\pi)^{p/2} \mathcal{I}_{p/2-1}(\kappa)}$$

is the regularizing constant to make it a probability density. We can observe that even though this distribution is supported on spherical $\mathcal{S}^{p-1}$, it is pretty heavy tailed. Take $p = 2$ as an example. It can be treated as a distribution of the angle $\theta$ between $\boldsymbol{x}$ and $\boldsymbol{\mu}$, with $\theta \in [-\pi, \pi)$. And we will have:

$$f_2(\theta; \boldsymbol{\mu}, \kappa) = C_2(\kappa) exp(\kappa \|\boldsymbol{\mu}\| \cos\theta)$$

[1] Corpus:https://drive.google.com/file/d/1fT1GxBMXEItf2NtNMjdXhA61HPecPO98/view

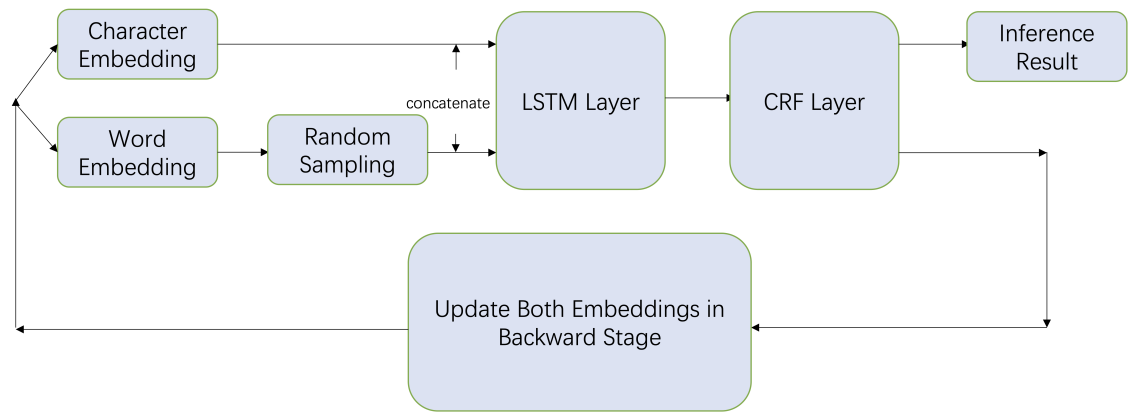

**Figure 4: Framework of Random Sampling Word Embedding**

And we will have that for $\theta = 0$ and $\theta = \pi$, the density at these two points is

$$f_2(0; \mu, \kappa) = C_2(\kappa) exp(\kappa \|\mu\|)$$

and

$$f_2(\pi; \mu, \kappa) = C_2(\kappa) exp(-\kappa \|\mu\|)$$

The larger the $\kappa$ is, the less heavy-tailed this distribution is. However, the $\kappa$ we obtain is relatively small. Thus it is still very likely that $\theta = \pi$, which means the sample is from the totally reverse section. A sample like this is certainly not a small noise or disturbance as we thought and certainly cannot make the results better. We have run experiments with this sampling distribution and we will compare the results.

To overcome this heavy tail problem, we propose to sample from the normal distribution, which is a pretty common idea to introduce disturbance. However, the problem with respect to this sampling is that we have to determine how to incorporate this $\kappa$. An intuitive idea is to model this $\kappa$ as the standard deviation of the random disturbance as following:

$$\boldsymbol{w}_i' = \boldsymbol{w}_i + \mathcal{N}(0, \kappa^2)$$

However since this $\kappa$ is not drawn from a normal distribution and cannot be directly linked to the variance, a hyper-parameter is introduced to scale the $\kappa$ and this hyper-parameter needs tuning in training. Thus the disturbance will be introduced as:

$$\boldsymbol{w}_i' = \boldsymbol{w}_i + \mathcal{N}(0, \kappa^2 / r^2) \tag{6}$$

Here we use $\boldsymbol{w_i}$ as original word embedding for word i and $\boldsymbol{w_i'}$ as our new embedding. $r$ is introduced as the regularizer to scale $\kappa$. The whole pipeline is presented in Figure 4. And the NER model will be built based on the work of Liu et al. [9]. Even though in NER model embeddings are fine-tuned along in the training process, the kappa will not be updated in the backward stage.

# 4 EXPERIMENT AND EVALUATION

In this section, we compare the result from the traditional NER model with traditional embedding and modified word embedding plus distribution specificity information.

## 4.1 Dataset

We have introduced the text corpus for training our own word embeddings in Section 3.2 with the link provided. And the dataset for training the NER model is the CoNLL03 dataset [18]. There exist label annotation mistakes by human annotators in the original datasets and [24] has provided a corrected test set which we adopt in this project. The train set, development set, and test set can be found here [2]. However, the sequence labeling method used in [24] is BIO which may not provide sufficient information to describe the exact location of a word in the text. Therefore, we convert the BIO labels into BIOES which includes Begin, Inside, Outside, End, and Single. We notice after using BIOES, the prediction accuracy and convergence speed are all improved.

## 4.2 Word Embedding Model

Our word embedding model is derived from Word2Vec model [15]. The main difference is the probability expression for $p(w_{i+j}|w_i)$ and training algorithm. They use $p\left(w_{i+j}|w_i\right) = \frac{\exp\left(\boldsymbol{u}_{w_i}^\top \boldsymbol{v}_{w_{i+j}}\right)}{\sum_{w' \in V} \exp\left(\boldsymbol{u}_{w_i}^\top \boldsymbol{v}_{w'}\right)}$ while we use equation (2). An extra $\kappa$ is added for each word in our model to incorporate distributional specificity information. The SGD is used to optimize the objective function in Word2Vec [15], but in this work, we use Riemannian SGD since we need to update parameters on a unit sphere [14]. The Skip-gram model and negative sampling method is adopted. We generate 5 negative samples for each positive sample and words with frequency lower than 5 will be dropped. Besides, the local context window size is set to 5. Our trained vMF-based embedding and $\kappa$ are available here [3].

## 4.3 Named Entity Recognition Model

The NER model used in this paper is a vanilla Char-LSTM-CRF model in [9]. This method employs pre-trained word embeddings and human annotations. It uses the word embedding as the input and supplements it with character embedding starting from random samples. Both the character embedding and the word embedding will be updated during the training process with all parameters in the model.

To be more specific, the NER model consists of several modules. As indicated in Figure 2 and Figure 3, there are two LSTM neural networks for forward word embedding and backward word embedding respectively. Then the character embedding will be concatenated with the word embedding and sent to another bi-directional LSTM neural network and the resulted embedding is loaded in a conditional random field. The output is fed into a decoder using Viterbi decoding algorithm to produce a probability distribution over all entities. The hyper-parameters used in the experiments are listed in Table 1.

## 4.4 Result Analysis

*4.4.1 Word Embedding Test.* In order to check the quality of our word embedding, we conduct word similarity test on the following datasets: WordSim353 [5], MEN [3] and SimLex999 [6]. Three different kinds of embedding including GloVe, Gauss-based embedding, and vMF-based embedding are trained on the same corpus. The test result of spearman coefficient is shown in Table 2. The results show that training word embedding under vMF distribution can achieve superior performance on the word similarity task. It is claimed that training the words as Gaussian distributions [23] can also capture the word similarity very well and outperform traditional embedding methods, but some important hyper-parameters are missing in the source code[4] and we can't reproduce the result after trying dozens of hyper-parameters.

[2] Corrected CoNLL03:https://github.com/ZihanWangKi/CrossWeigh/tree/master/data
[3] Result:https://drive.google.com/drive/u/0/folders/1s8Y3hZ68r2p4vOJc8XV4BmCupL_JMdHB

**Table 1: Hyper-Parameters in NER**

| | |
|---|---|
| Drop Rate | 0.5 |
| Optimizer | SGD |
| Learning Rate | 0.015 |
| Learning Rate Decay Rate | 0.05 |
| Word Embedding Dimension | 100 |
| Character Embedding Dimension | 50/30 |

**Table 2: Word Similarity Result of Three Types of Embedding**

| Embedding | WordSim353 | MEN | SimLex999 |
|---|---|---|---|
| GloVe | 0.539 | 0.666 | 0.296 |
| Gauss-based | 0.499 | 0.581 | 0.211 |
| vMF-based | 0.706 | 0.726 | 0.319 |

**Table 3: An Example of Word Analogy Test**

| Relation type | Pair 1 | | Pair 2 | |
|---|---|---|---|---|
| gram2-opposite | acceptable | unacceptable | aware | unaware |
| gram4-superlative | bad | worst | big | biggest |
| capital-common-countries | Athens | Greece | Baghdad | Iraq |
| capital-world | Abuja | Nigeria | Asmara | Eritrea |
| family | boy | girl | groom | bride |

Then the performance of word analogy is evaluated on a comprehensive test set containing nine types of syntactic questions, and five types of semantic questions [15]. Some examples of the questions are shown in Table 3. Given the information of the first three words, the last word should be predicted correctly by performing simple algebraic operations with the word embedding. One example is shown in Figure 5. Table 4 concludes the result of word analogy test. The vMF-based word embedding can beat GloVe embedding while the Gauss-based embedding fails again. It can be seen from the above result that the quality of vMF-based distribution is quite good so we use this embedding along with its distributional specificity information to conduct experiments in the NER model.

*4.4.2 NER Model Performance.* Using GloVe embedding and Gauss-based embedding, we can get averaged performance 90.61 and 88.07

[4]Source code: https://github.com/seomoz/word2gauss

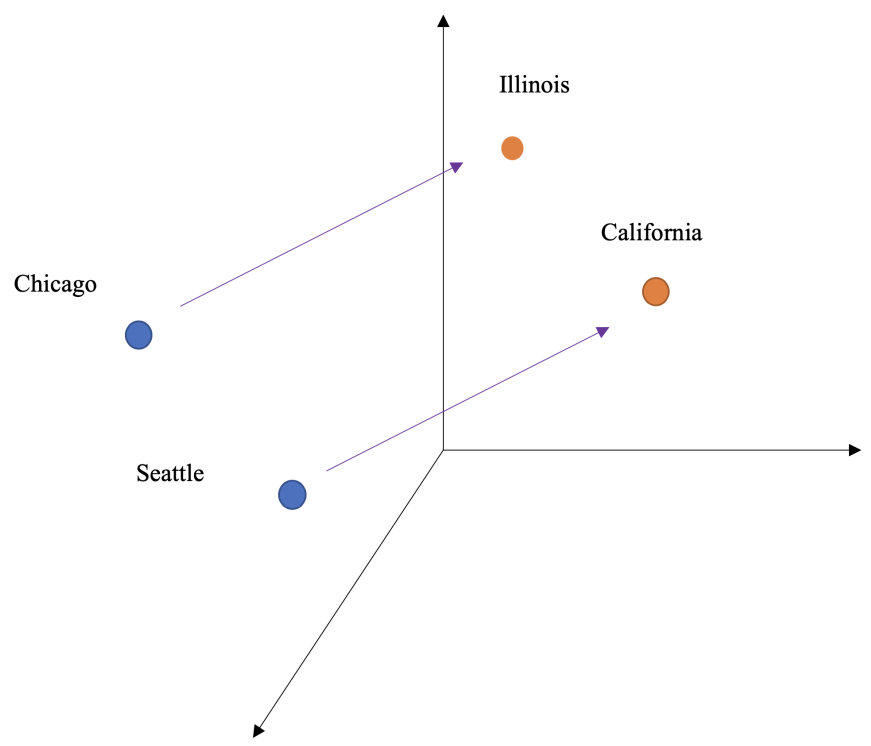

**Figure 5: An Example of Word Analogy**

**Table 4: Word Analogy Result of Three Types of Embedding**

| Embedding | Syntactic accuracy(%) | Semantic accuracy(%) |
|---|---|---|
| GloVe | 52.68 | 49.58 |
| Gauss-based | 18.22 | 23.79 |
| vMF-based | 58.11 | 59.36 |

**Table 5: Result from Concatenated Word Embedding**

| Embedding | F1 Score |
|---|---|
| GloVe | 90.61 (±0.14) |
| Gauss-based embedding | 88.07 (±0.12) |
| vMF-based embedding | 90.95 (±0.27) |
| vMF-based embedding $\oplus\ \kappa$ | 90.78 (±0.09) |
| vMF-based embedding $\oplus\ \boldsymbol{\kappa}$, $\boldsymbol{\kappa} \in \mathcal{R}^{10}$ | 90.96 (±0.22) |

respectively. We then try three different methods as described above on vMF-based embedding.

The result of method 1 is summarized in Table 5, in which the F1 scores are averaged from 5 runs and the standard deviation is reported. The results validate the effectiveness of vMF-based embedding since the three results obtained from it are all better than the results from GloVe embedding and Gauss-based embedding. And by using the concatenated vMF-based embedding with $\boldsymbol{\kappa} \in \mathcal{R}^{10}$, we achieve the highest mean value of F1 score. However, the performance of word embedding concatenated with a single value of $\kappa$ is not as good. The reason is that the dropout rate in the NER model is high at 0.5, the information provided by a single value may not be recognized compared with the original 100-dim word embedding vector. It is also worth noticing that the performance of concatenated word embedding with one single $\kappa$ value has a smaller standard deviation which guarantees a more reliable result in other downstream tasks compared with the other two vMF-based results.

For method 2 which dynamically adjusts composition between word embedding and character embedding, the averaged F1 score is 90.70. The performance is slightly better than GloVe embedding, but there is no improvement after adding $\kappa$ as a signal in the selection gate.

**Table 6: Result from Embedding with Random Sampling**

| Method | F1 Score |
|---|---|
| Sampling from vMF | 89.92 (±0.08) |
| Sampling from normal with Kappa being variance | 90.53 (±0.22) |
| Sampling from normal with Kappa being std and $r = 40$ | 90.50 (±0.31) |
| Sampling from normal with Kappa being std and $r = 10$ | 90.49 (±0.05) |
| Sampling from normal with Kappa being std and $r = 5$ | 90.75 (±0.16) |
| Sampling from normal with Kappa being std and $r = 2$ | 89.84 (±0.27) |

**Table 7: Performance Comparison**

| Method | F1 Score |
|---|---|
| GloVe | 90.61 (±0.14) |
| vMF-based embedding | 90.95 (±0.27) |
| Method 1 | 90.96 (±0.22) |
| Method 2 | 90.70 (±0.16) |
| Method 3 | 90.75 (±0.16) |

The result of Method 3 is summarized in Table 6. Here $r$ is the re-scaling hyper-parameter as indicated in (6). All scores are averaged over 3 experiments and the standard deviation is reported. It can be observed that with the noise being sampled from vMF distribution, the F1 score drops severely, corroborating the argument that the heavy tail of the vMF distribution may introduce too much noise in the sampling, thus resulting in a worse performance of the model. Even though the introduction of the normal noise is not improving the F1 score to a large extent, it can be observed that with proper tuning of the re-scaling hyper-parameter, which in our case is $r = 5, 10$, the result will be less volatile, which indicates a more robust model is introduced in the sense that the performance has a less standard deviation. If the randomness is too small, which in our case is $r = 40$, the standard deviation is not reduced too much. On the contrary, if the randomness is too large, which in our case is $r = 2$, the standard deviation increases, due to a relatively large disturbance.

## 5 CONCLUSION AND FUTURE WORK

In this work, we mainly focus on how to incorporate the distribution information of the embedding, specifically variance of the embedding, into the NER framework. We train a new kind of embedding under vMF distribution and develop three main methods: concatenating the variance to the original embedding, using variance as a dynamic signal for feature selection, and random disturbance added to embedding. Finally, we find that directly using distribution-aware embedding alone in the model can get reasonable improvement compared with the benchmark. And adding $\kappa$ in the above three methods also shows slight enhancement. What is more important is that two methods can reduce the volatility of the performance and rendering a more robust model.

However, we don't observe statistically significant improvement in F1 score after adding $\kappa$ to the NER model compared with bare vMF-based embedding. One possible reason may be that the dimension of $\kappa$ is negligible compared with word embedding, the specificity information is highly compressed thus it may be covered by the word embedding and the model can't successfully decode it. So we plan to train $\kappa$ as a vector for each dimension of word embedding instead of the whole word embedding. By doing this, the specificity information could be delivered to the NER model in a more straightforward way.

To explore the methods further, one possible future work will be adopting the word shape idea from [22], in which we may train a separate word embedding for different value bins of $\kappa$ where words belonging to the same $\kappa$ level are encouraged to appear closer to each other and vise versa. We may also emphasize other methods incorporating variance in the NER framework, for example using the variance as a possibility to stop fine-tuning the embedding and return to pre-trained embedding. Another possible track is to extend the idea of distribution aware embedding to other downstream tasks, such as Question Answering.

## 6 ACKNOWLEDGEMENTS

This work is supported by the Data Mining Group (DMG) of the Computer Science Department at the University of Illinois at Urbana-Champaign. We thank Liyuan Liu for valuable guidance. We also thank Professor Jiawei Han and all teaching assistants for constructive comments and feedback.